\documentclass{article}

\PassOptionsToPackage{numbers}{natbib}
\usepackage[preprint]{neurips_2023}

\usepackage[utf8]{inputenc} 
\usepackage[T1]{fontenc}    
\usepackage{hyperref}       
\usepackage{url}            
\usepackage{booktabs}       
\usepackage{amsfonts}       
\usepackage{nicefrac}       
\usepackage{microtype}      
\usepackage{xcolor}         
\usepackage{subfig}
\usepackage{graphicx}
\usepackage{amsmath}

\title{ShapeShift: Superquadric-based Object Pose Estimation for Robotic Grasping}

\author{%
  E. Zhixuan Zeng, Yuhao Chen, Alexander Wong\\
  Vision and Image Processing Research Group, University of Waterloo, Waterloo, Canada \\
  \texttt{\{emilyzhixuan.zeng, yuhao.chen1, alexander.wong\}@uwaterloo.ca} 
}

\begin{document}

\maketitle

\begin{abstract}
  Object pose estimation is a critical task in robotics for precise object manipulation.  However, current techniques heavily rely on a reference 3D object, limiting their generalizability and making it expensive to expand to new object categories. Direct pose predictions also provide limited information for robotic grasping without referencing the 3D model. Keypoint-based methods offer intrinsic descriptiveness without relying on an exact 3D model, but they may lack consistency and accuracy. To address these challenges, this paper proposes ShapeShift, a superquadric-based framework for object pose estimation that predicts the object's pose relative to a primitive shape which is fitted to the object. The proposed framework offers intrinsic descriptiveness and the ability to generalize to arbitrary geometric shapes beyond the training set.
\end{abstract}

\section{Introduction}
\label{sec:intro}


    Object pose estimation is a crucial task in robotics, enabling precise manipulation of objects in the environment. However, a common challenge faced by current object pose estimation techniques is their heavy reliance on a reference 3D object.  
    Adding new object categories into the model incurs a substantial expense, as it necessitates accurate 3D scans.
    Furthermore, these models have limited generalizability and are confined to a small set of objects. In practice, referring back to the 3D model to obtain useful information about grasp points and positions can limit the effectiveness of these techniques in robotic grasping applications.



    Keypoint-based methods for object pose estimation, such as those described in \cite{robson2022keypoint_grasp} offer intrinsic descriptiveness about the 3D object without the need to reference a 3D model. However, these methods can be arbitrary in their definition and lack consistency between objects. Such methods deal poorly with multiple degrees of symmetry and heavy occlusion.


    To overcome these challenges, this paper proposes ShapeShift, a framework for object pose estimation based on primitive shapes. By fitting a primitive shape to an object, the proposed approach provides intrinsic descriptiveness and information about the 3D object without relying on a 3D model. In this paper, we specifically utilize superquadrics, which are three-dimensional shapes described by a mathematical equation that has been used to simplify shape representation in previous works \cite{liu2022robust, paschalidou2020hierarchical_superquad, Oblak2021expli_imp_superq}. The proposed approach predicts the object's pose in reference to a predicted primitive shape fitted to the object. This not only provides intrinsic descriptiveness but also enables generalization to arbitrary geometric shapes not present in the training set, making it a promising solution to the challenges faced by current object pose estimation techniques.

\begin{figure}
    \centering
    \includegraphics[width=0.45\linewidth,trim={3cm 4cm 2cm 6cm},clip]{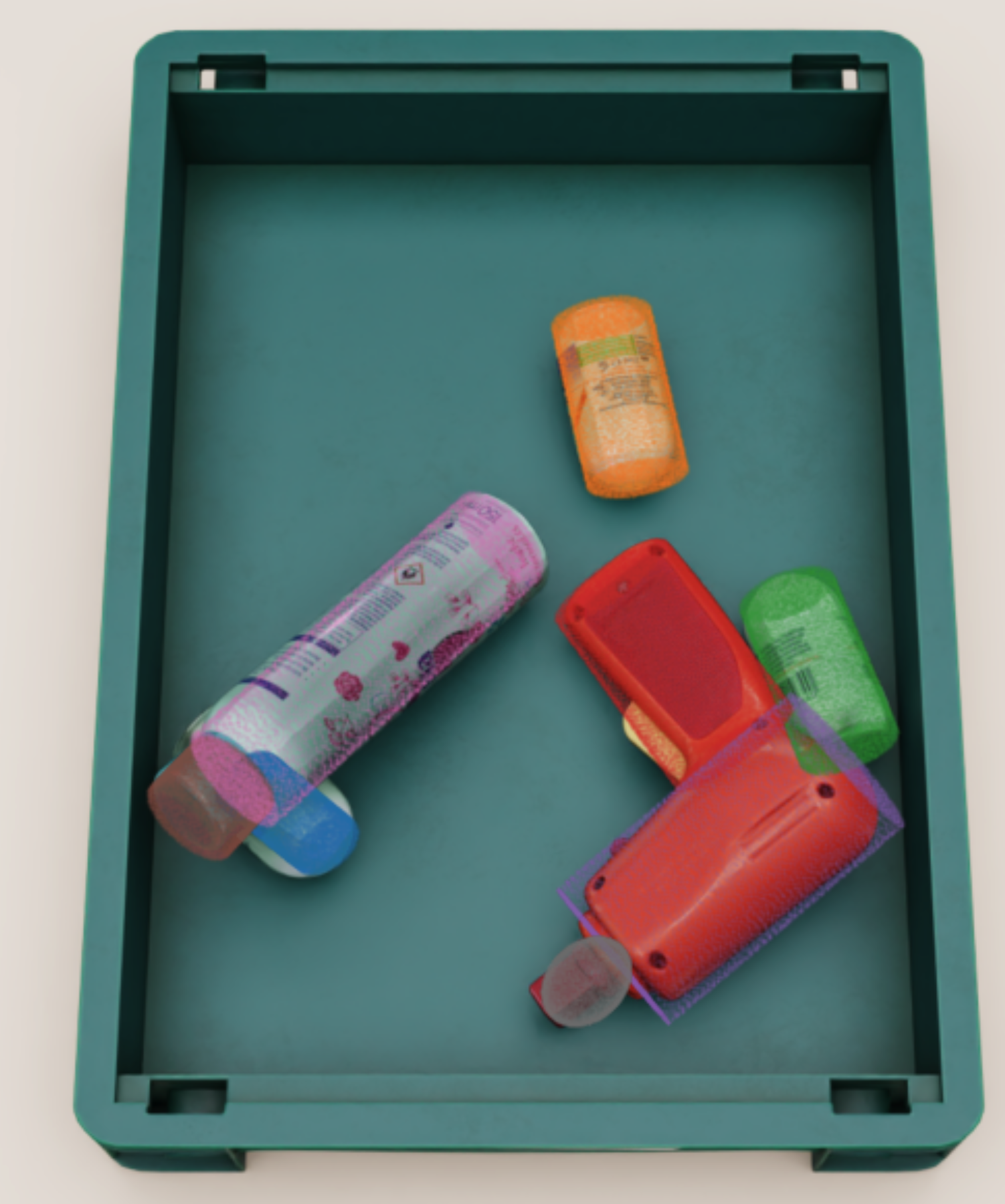}
    \includegraphics[width=0.45\linewidth]{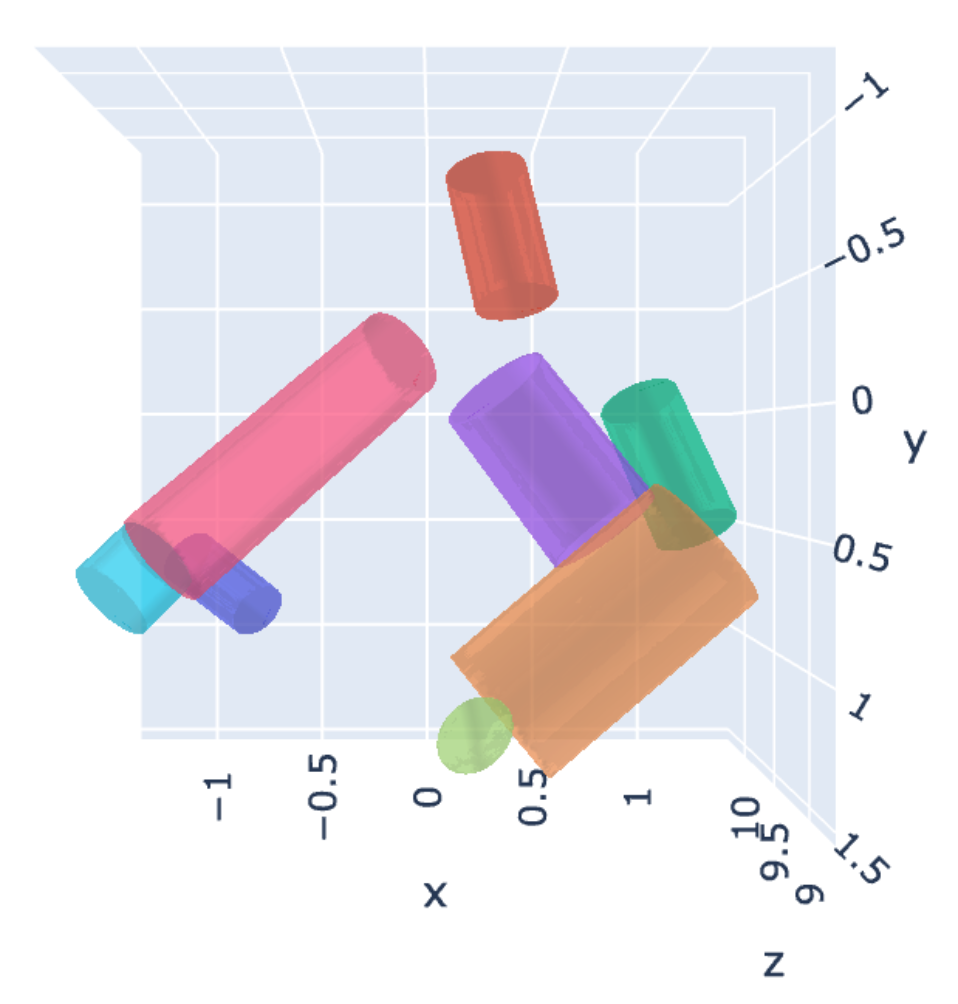}
    \caption{An example of ShapeShift on a typical scene.}
    \label{fig:scene_example}
\end{figure}

\section{Method}

The proposed ShapeShift framework can be described as follows. In the first phase, primitive shapes are fitted to each geometric part of an object using superquadrics.  In the second phase, the superquadrics fits are leveraged as ``ground truth`` for a superquadric-guided direct regression network to directly predict pose and shape information. 

\subsection{Phase 1: Superquadric fitting}
Superquadrics,
characterized by $\theta = \left\{ \epsilon_1, \epsilon_2, a_x, a_y, a_z, [r_x, r_y, r_z], t \right\}$. $a_x$, $a_y$, and $a_z$ are the scale, $\epsilon_1$ and $\epsilon_2$ define the shape of the surface, ($r_x, r_y, r_z$) defines rotation and $t$ is translation. 




The first phase of ShapeShift leverages the superquadric fitting technique from Liu \cite{liu2022robust} to fit superquadrics to parts. 


However, there are scenarios where different parameters result in the same superquadric. In particular, Liu \cite{liu2022robust} introduced duality similarity, when $a_x \approx a_y$, $\epsilon_2 > 1$:

    
\begin{equation}
        \theta_3^c = \left\{ \epsilon_1, 2-\epsilon_2, s\cdot \bar{a}, s \cdot \bar{a}, a_z, R\cdot R_z(\pi/4), t \right\} \\
    \label{eq:duality}
\end{equation}
where
\begin{equation}
    \begin{split}  
        s &= (\sqrt2/2-1)\epsilon_2+2-\sqrt2/2 \\ 
        \bar{a} &= (a_x+a_y)/2
    \end{split}
\end{equation}





To avoid such cases and maintain $\epsilon_2$ within the range $[0, 1]$, the scale can be redefined using a warp transformation. Additionally, the rotation can be transformed into a new rotation using the following approach:

\begin{equation}
\begin{split}
    S &= \begin{bmatrix} a_x*s, &  a_y*s,  & a_z\end{bmatrix}^T \mathbf{I} \\
    S_{warp} &= R_d^{-1}SR_d \\
    R_{new} &= R R_d
\end{split}
\end{equation}


\noindent where $R_d = R_z(\pi/4)$ is a rotation of $\pi/4$ about the z axis

Instead of a 3x3 transformation matrix for S, it can be further expanded to a pure scale and a shear transformation. Reversely, $S$ and $R$ can be combined into a single 3x3 transformation matrix ($S_{}$). 

\subsection{Phase 2: Superquadric-guided Pose Estimation}
\begin{figure}
    \centering
    \includegraphics[width=\linewidth]{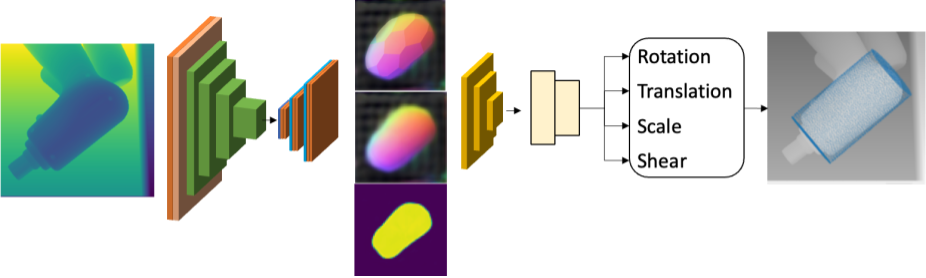}
    \caption{The proposed superquadric-guided pose estimation architecture extends upon~\cite{wang2021gdr} by introducing a new scale and shear head. The main difference is defining the 3D correspondences in reference to a superquadric shape rather than the 3D model of the full object.}
    \label{fig:architecture}
\end{figure}

The second phase of ShapeShift involves superquadric-guided pose estimation to directly predict pose and shape. The proposed architecture (see Figure \ref{fig:architecture}) uses depth images as input and extends upon~\cite{wang2021gdr} in several ways. To reduce search dimensions, superquadrics were discretized based on shape parameters and treated as different object categories. This also allows for easily defining symmetry (discrete and continuous) for error calculations. To generate an intermediate representation, un-scaled primitive shapes were sampled using furthest point sampling.  Additionally, the proposed architecture introduces additional scale and shear heads. An additional head can be added in the future to predict shape parameter offsets.


\begin{figure}
    \centering
    \subfloat[\label{fig:superquadrics_space}]{
        \includegraphics[width=0.35\linewidth]{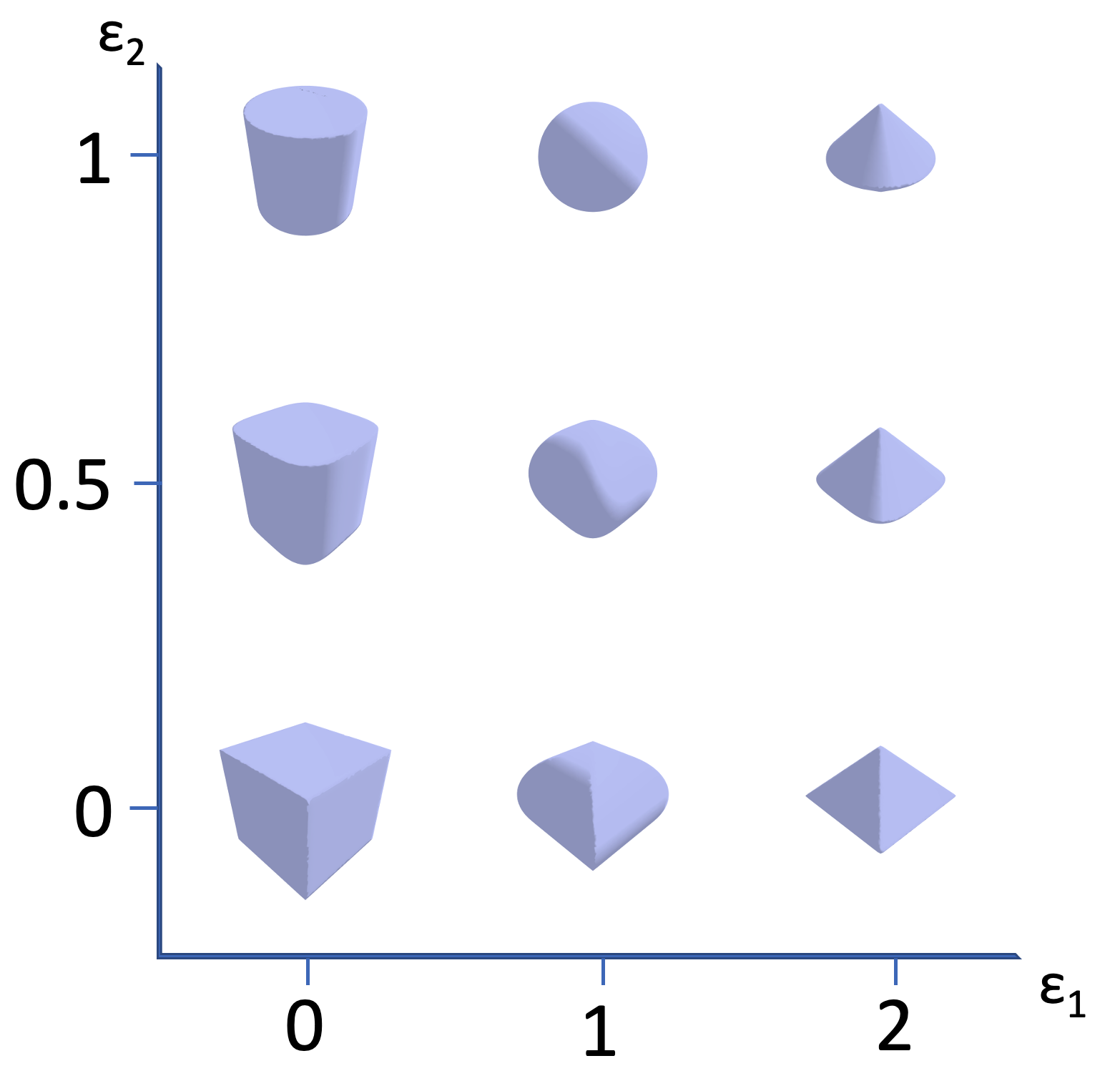}
    } 
    \subfloat[\label{fig:clutter}]{
        \includegraphics[width=0.35\linewidth,trim={0cm 4cm 0cm 1cm},clip]{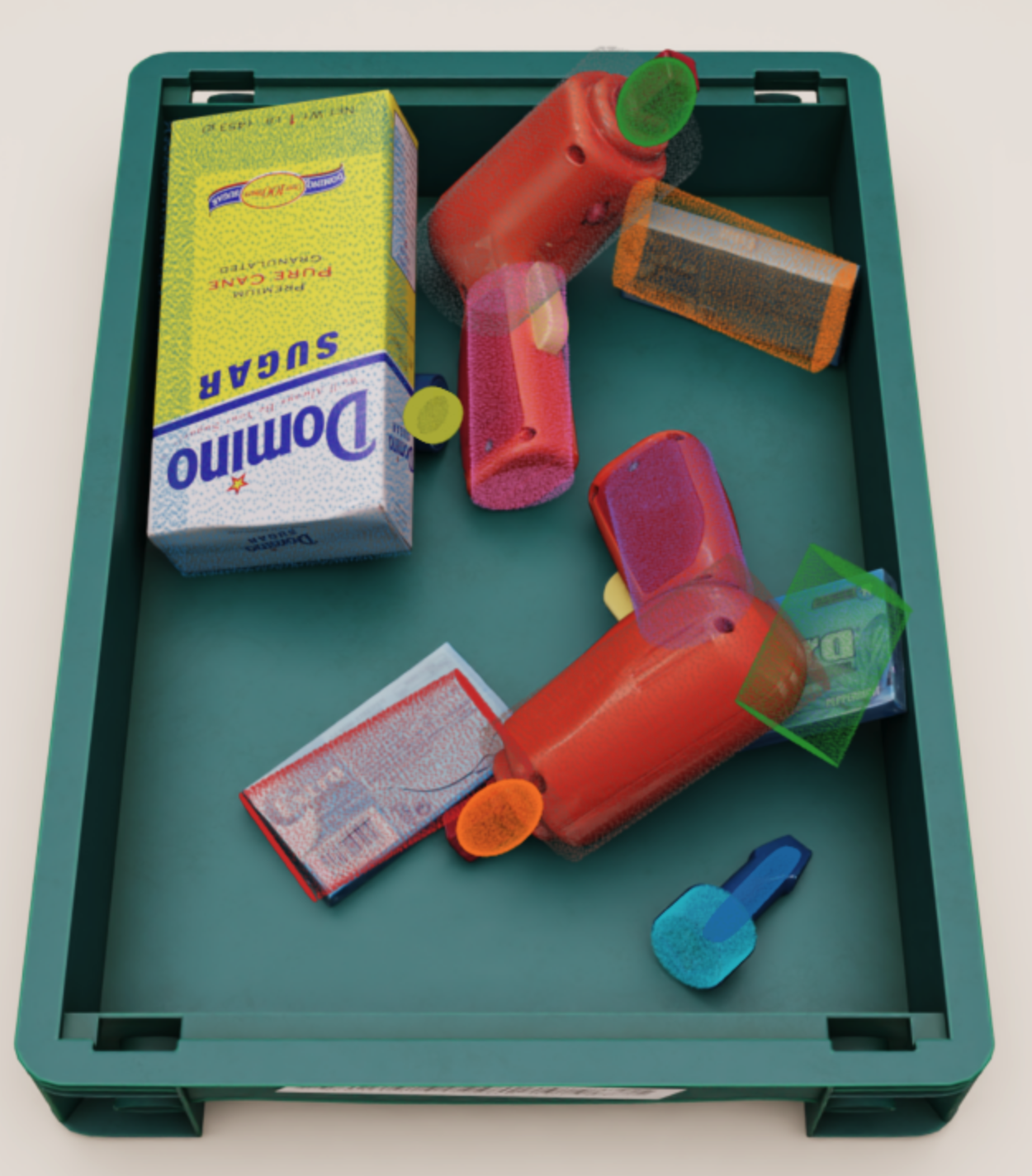}
    } \\
    \subfloat[\label{fig:occlusion_1}]{
        \includegraphics[width=.35\linewidth,trim={4cm 10cm 2cm 2cm},clip]{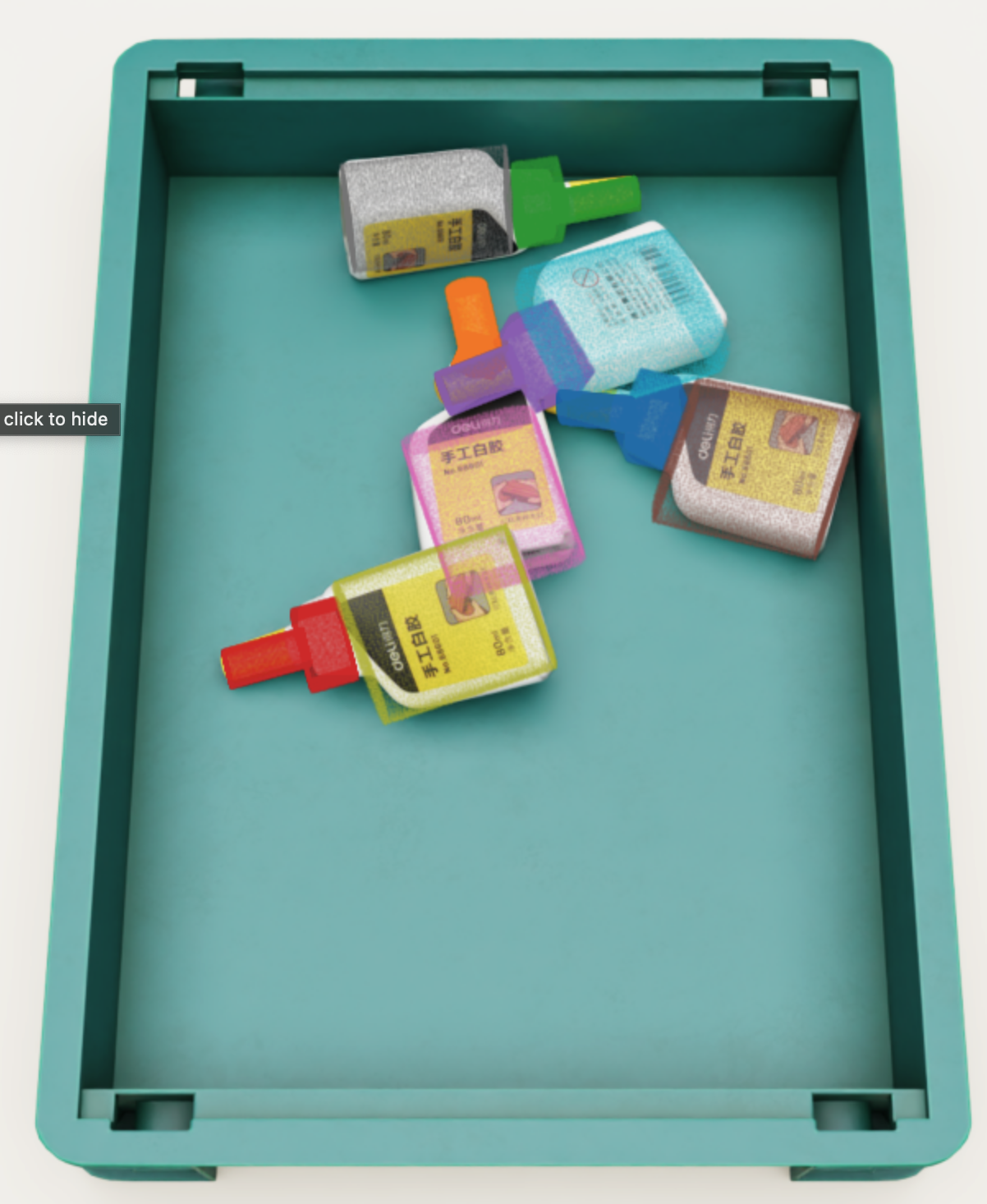}
    }
    \subfloat[\label{fig:occlusion_2}]{
        \includegraphics[width=.35\linewidth,trim={4cm 10cm 5cm 0cm},clip]{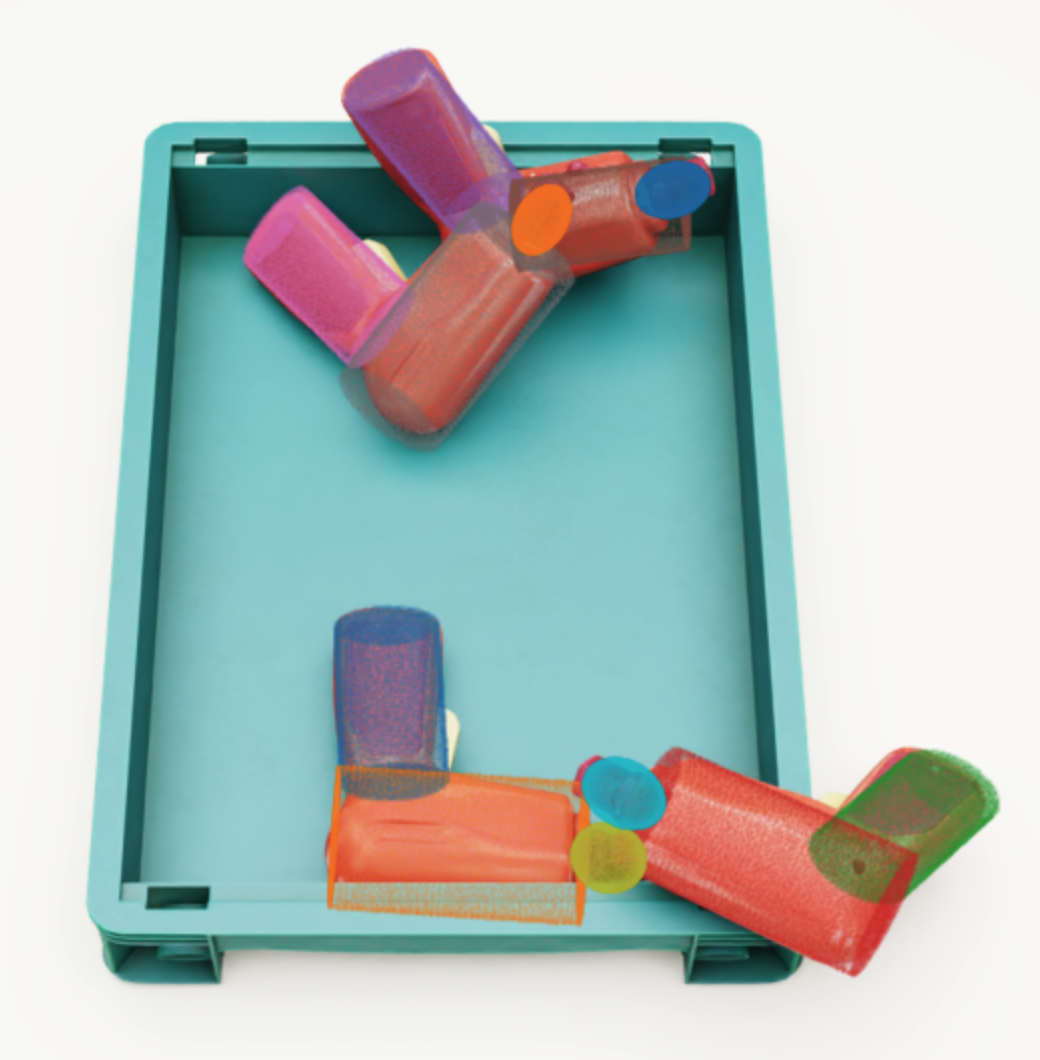}
    }
    \caption{Superquadrics shape space and prediction examples}
    \label{fig:examples}
    
\end{figure}

\begin{figure}
    \centering
    \includegraphics[width=.49\linewidth,trim={0cm 0cm .4cm 0cm}]{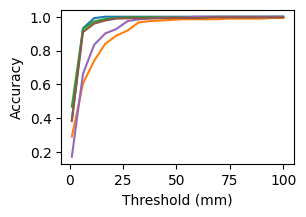}
    \includegraphics[width=.49\linewidth,trim={3cm 0cm 0cm 0cm}]{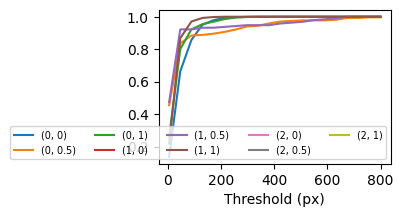}
    \caption{MSSD (left) and MSPD (right) based accuracy scores over different  thresholds on discrete superquadric shapes}
    
    \label{fig:accuracy}
\end{figure}

\section{Experimental results and discussion}

We conducted experiments on a subset of the MetaGraspNet benchmark dataset \cite{chen2021metagraspnet} containing 54 object categories of objects that are composable using multiple superquadrics. We evaluated the performance of our proposed ShapeShift framework for object pose estimation using the Maximum Symmetry-Aware Surface Distance (MSSD) and Maximum Symmetry-Aware Projection Distance (MSPD) metrics \cite{hodavn2020bop}. For evaluation, we combined rotation, scale, and shear into a single transformation matrix. The accuracy scores based on both metrics are shown in Figure \ref{fig:accuracy}.

During experimentation, we observed three challenges. 

First, the method's performance was lower for underrepresented shape categories, as shown in Figure \ref{fig:accuracy}. 

Second, shear is underrepresented in the ground truth, but there exists a head dedicated to predicting it, causing a number of false predictions. False predictions of shear caused projected error to be higher than 3D surface distance error for $\epsilon = (0, 0)$ shapes. 


Finally, the proposed method is robust to partial occlusion on objects composed of multiple primitive shapes. As illustrated in Figures \ref{fig:occlusion_1}, \ref{fig:occlusion_2}, parts of the object that are not occluded can still be properly predicted, even while the method struggles on heavy occlusion. Further work exploring data augmentation methods and an addition amodal mask prediction head should help with overall occluded object performance.

\subsection{Conclusion and future work}

In this paper, we introduced ShapeShift, a superquadric-based framework for object pose estimation is proposed that predicts an object's pose relative to a primitive shape fitted to the object. This approach provides intrinsic descriptiveness and information about the 3D object without relying on a 3D model. The approach was further tested on the MetaGraspNet benchmark dataset, and has demonstrated the ability to approximate the shapes present in the image.
Future work would focus on optimizing performance in cases of occlusion, dealing with the imbalanced distribution of shape types, adding an additional head to predict precise shape parameters, and evaluating performance on novel objects not seen in the training set.

{\small
\bibliographystyle{plainnat}
\bibliography{egbib}
}

\end{document}